\documentclass{article}

\usepackage[preprint]{corl_2026} 
\usepackage{amsmath,amsfonts,amssymb}
\usepackage{algorithmic}
\usepackage{algorithm}
\usepackage{array}
\usepackage{bbding}
\usepackage{color}
\usepackage{booktabs}
\usepackage{threeparttable}
\usepackage{textcomp}
\usepackage{stfloats}
\usepackage{url}
\usepackage{verbatim}
\usepackage{graphicx}
\usepackage{cite}
\usepackage{natbib}
\usepackage[caption=false,font=normalsize,labelfont=sf,textfont=sf]{subfig}

\newcommand{\name}{AeroAct}

\title{\textsc{\name}: Action-Centered World-Action Models for Language-Conditioned Quadrotor Flight}

\author{
  Xinhong Zhang$^{1,*}$, Qiyuan Zhu$^{1,*}$, Yubo Huang$^{1,*}$, Haolin Chen$^{2,*}$, Runqing Wang$^{1}$, \\
  \textbf{Yuhao Mo$^{1}$, Zhongxin Chen$^{1}$, Yu Hu$^{3}$, Xinjiang Wang$^{3}$, Jian Sun$^{1}$, Gang Wang$^{1,\dagger}$}
  \vspace{0.2cm}
\\
  $^{1}$School of Automation, Beijing Institute of Technology\\
  $^{2}$School of Mechanical Engineering, Beijing Institute of Technology\\
  $^{3}$Independent researcher\\
  {\footnotesize $^{*}$ Equal contribution \quad $^\dagger$ Corresponding author}
}

\begin{document}
\maketitle


\allowdisplaybreaks

\begin{figure*}[!h]
\centering
\vspace{-1cm}
\includegraphics[width=0.96\textwidth]{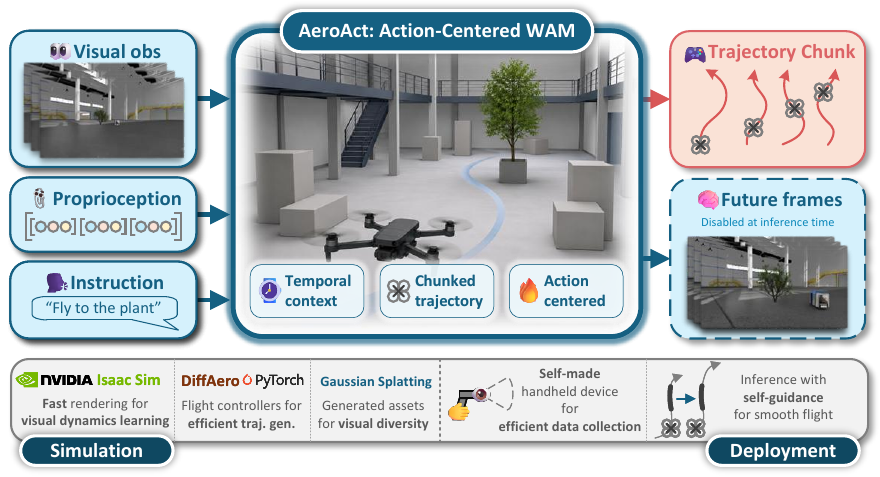}
\vspace{-0.cm}
\caption{\textbf{Overview of \name.} The policy receives egocentric visual history, proprioception and language, and predicts a chunked local trajectory for aerial navigation. Training uses paired flight actions and future visual observations from Isaac, DiffAero, 3DGS scenes, and real-world handheld data; deployment decodes actions with self-guidance and executes these actions through a trajectory controller.}\label{fig:cover}
\vspace{-0.cm}
\end{figure*}

\begin{abstract}

Language-conditioned quadrotor flight requires a policy to ground semantic goals, anticipate the visual consequences of ego-motion, and output control references that remain smooth and dynamically executable under rapidly changing first-person views. Existing aerial vision-language navigation and vision-language-action methods commonly use discrete actions, high-level waypoints, or instantaneous velocity commands, which provide limited supervision about how flight actions change future observations. 
We present \name{}, an action-centered world-action model (WAM) for quadrotor navigation. To the best of our knowledge, \name{} is the first WAM instantiated and demonstrated for real-world aerial flight. The model adapts a pretrained video diffusion Transformer to predict local trajectory-action chunks from egocentric visual history, proprioception, and language. Future first-person frames are used during training as dense consequence supervision, while deployment directly decodes actions without generating future video. To obtain aligned visual, state, language, and dynamically feasible action data, we build a DiffAero-based pipeline with complementary Isaac Lab and 3D Gaussian splatting renderers. We further introduce a low-cost handheld collection device that couples camera observations with motion estimates to recreate flight-like egocentric trajectories, and a self-guidance procedure that improves temporal consistency across overlapping trajectory chunks. Closed-loop simulation and real-world experiments show that temporal visual context improves target tracking and object-search performance, and that WAM-based policies can be executed on a physical quadrotor.

\end{abstract}


\section{Introduction}
\label{sec:intro}

Aerial robots are increasingly expected to execute natural-language commands while flying safely through cluttered 3D scenes. This problem is substantially more demanding than standard vision-language navigation (VLN). The robot observes the world through a rapidly moving egocentric camera, its actions immediately alter the future visual stream, and unsafe commands can quickly lead to oscillation, loss of target visibility, or collision. A practical language-conditioned flight policy must therefore do more than identify the goal in an image: it must connect semantic grounding with a control representation that is local, smooth, and compatible with quadrotor dynamics.

Learning-based agile flight has shown that reinforcement learning and imitation learning can produce highly reactive policies for racing, obstacle avoidance, target tracking, and aggressive maneuvering. These systems demonstrate the effectiveness of end-to-end perception and control, but they are often trained with task-specific rewards or expert demonstrations and typically condition on geometric goals rather than open-vocabulary instructions. In parallel, vision-language-action (VLA) models have improved semantic generalization by attaching action heads to large vision-language representations. However, direct action decoding alone does not explicitly teach the policy how first-person observations evolve under its own actions. Recent world-model advances, including Dreamer-style latent dynamics and generative interactive simulators, suggest that predicting future observations can provide a compact substrate for control, planning, and representation learning~\citep{dreamerv3,storm,genie,unisim,romero2025dream,zhang2025corb,zhang2026mad,feng2026multi}. World-action models (WAMs) push this idea further by jointly training visual consequence prediction and action generation, allowing dense video supervision to regularize the learned action distribution~\citep{gwp,dreamzero}.

Bringing WAMs to quadrotor navigation is nevertheless nontrivial. First, training a flight WAM requires large-scale multimodal trajectories with tightly synchronized egocentric video, language instructions, proprioceptive states, and dynamically feasible flight actions. Such data are much harder to collect than tabletop demonstrations because quadrotor motion is fast, safety-critical, and constrained by underactuated dynamics. Second, video-centric WAMs can be expensive at inference time. Autoregressive future-video generation increases latency, and small visual prediction errors may accumulate and corrupt downstream action decoding. These failure modes are particularly problematic for aerial robots, whose low-level controllers require smooth, locally executable trajectory references to maintain stable flight.

We address these challenges with \name{}, an action-centered WAM for language-conditioned quadrotor flight. Instead of generating future video during deployment, \name{} uses a history of egocentric observations and proprioceptive states to infer the current flight context and predicts a chunk of local fifth-order trajectory parameters. Future visual prediction is retained only as dense auxiliary supervision during training. A blockwise causal mask prevents the action stream from accessing future-frame tokens, so that deployment can decode actions autoregressively without test-time video generation. This design preserves the temporal and geometric priors of video models, reduces inference-time computation by $37.8$\%, and avoids error accumulation from imagined frames.

To address the data bottleneck, we construct a scalable hybrid pipeline that combines DiffAero dynamics~\citep{diffaero}, Isaac Lab, and 3D Gaussian splatting (3DGS)~\citep{3dgs,jia2025discoverse,jia2026gsplayground}. The pipeline produces aligned egocentric video, language, proprioception, and dynamically feasible action chunks across tracking and object-reaching scenarios. We also introduce a handheld collection device that approximates flight-like egocentric motion, enabling low-cost real-world fine-tuning without requiring extensive autonomous flight data collection. The experiments validate the design choices: longer temporal context substantially improves target tracking and object search, mixed-domain training maintains strong closed-loop performance, and real-world flight demonstrates that the learned action interface can be executed on a physical quadrotor.

The main contributions of this paper are as follows:
\begin{itemize}
    \item We introduce \name{}, an action-centered, history-aware WAM for language-conditioned quadrotor flight. Future-frame prediction provides dense consequence supervision during training, while deployment decodes smooth trajectory-action chunks without inference-time video generation.
    \item We develop a scalable data-generation pipeline that aligns egocentric RGB, proprioception, language, and dynamically feasible actions using DiffAero with Isaac Lab and 3DGS rendering, together with a handheld real-world collection device.
    \item We formulate a smooth chunked action space based on local fifth-order trajectory segments, introduce inference-time self-guidance for temporal consistency, and demonstrate, to the best of our knowledge, the first real-world quadrotor flight using a WAM-based policy.
\end{itemize}

\section{Related Work}
\label{sec:relatedwork}
\subsection{Aerial Vision-Language Navigation}

Aerial VLN asks a quadrotor to follow language instructions in three-dimensional environments from egocentric observations. Early benchmarks established the task: AVDN introduced a dialog-based commander-agent setting~\citep{advn}, and AerialVLN provided city-scale simulation with human-annotated routes~\citep{aerialvln}, highlighting the importance of spatial relations, temporal ordering, and directional references in aerial scenes. Later platforms improved realism and scale. OpenUAV emphasizes realistic target-oriented flight~\citep{openuav}; OpenFly integrates multiple rendering engines and an automated toolchain for large-scale data generation~\citep{openfly}; and UAV-Flow reframes aerial navigation as fine-grained ``flying-on-a-word'' imitation learning~\citep{uavflow}. A common limitation is that many action spaces are inherited from ground VLN, such as discrete moves or high-level waypoints, which abstract away the continuous, dynamics-coupled behavior of quadrotors.

Recent methods use pretrained vision-language models for zero-shot aerial navigation, such as OnFly~\citep{onfly}, or train VLA policies for onboard deployment, such as VLA-AN~\citep{vlaan}. These approaches improve semantic grounding and long-horizon instruction following, but their action interfaces are often waypoints, primitives, or velocity commands. As a result, the policy may select semantically plausible actions without explicitly learning their visual and dynamical consequences. \name{} complements this line by coupling trajectory-level action chunks with future first-person observations during training, while retaining a command representation suitable for quadrotor control.

\subsection{World-Action Models}

World-action models, recently formalized by \citet{dreamzero}, are robot policies built on video-generation backbones that jointly model future observations and actions. They differ from classical world models, such as Dreamer, Genie, and UniSim~\citep{dreamerv3,genie,unisim}, which learn predictive dynamics but do not by themselves define a deployable closed-loop policy. They also differ from standard VLA policies, which map observations and language directly to actions but typically lack an explicit objective for modeling the visual consequences of actions. By co-training video and action prediction, WAMs use video as dense supervision for physical dynamics and transfer temporal-geometric priors from the video backbone to the action stream.

Recent WAMs differ in backbone and action-generation design. Action-video WMs couple video and action diffusion on large-scale robot datasets~\citep{uwm}; Cosmos Policy fine-tunes a video model for visuomotor control~\citep{cosmospolicy}; DreamZero introduces autoregressive joint denoising with closed-loop caching~\citep{dreamzero}; Motus uses separate understanding, video-generation, and action experts~\citep{motus}; Fast-WAM avoids future-frame generation during inference~\citep{fastwam}; and GigaWorld-Policy pursues an action-centered design for efficient control~\citep{gwp}. These methods are mainly developed for tabletop manipulation, where cameras are comparatively stable and actions parameterize end-effectors or joints. Aerial navigation is substantially different: the egocentric viewpoint changes rapidly, and commands must remain smooth, local, dynamically feasible, and safe. To the best of our knowledge, \name{} is the first action-centered WAM for quadrotor flight on a physical platform.

\section{Methodology}
\label{sec:methodology}
\subsection{Problem Formulation}

We formulate language-conditioned aerial navigation as sequential prediction from first-person sensing. At time $t$, the robot observes an RGB image $o_t$, a proprioceptive state $s_t$, and a language instruction $l$. Instead of acting from a single frame, \name{} conditions on a temporally strided observation history
\begin{equation}
  \mathcal O_t^h=\left(o_{t-(h-1)\Delta},o_{t-(h-2)\Delta},\ldots,o_t\right),
\end{equation}
where $h$ is the number of reference frames and $\Delta$ is the temporal stride. The policy predicts an action chunk $a_{t:t+p-1}=(a_t,a_{t+1},\ldots,a_{t+p-1})$ of length $p$. Each action compactly parameterizes a dynamically executable local trajectory segment, rather than an instantaneous discrete command or a global waypoint. Following the WAM formulation~\citep{gwp}, we use a unified model $g_\Theta$ for both action prediction and visual consequence modeling,
\begin{equation}
  \left(a_{t:t+p-1},c_t\right)\sim g_{\Theta}\left(\cdot|\mathcal O_t^h,s_t,l\right),
\end{equation}
where $c_t$ is an action latent used to condition visual forecasting. Given the same context and $c_t$, the model predicts future observations
\begin{equation}
\left(o_{t+\Delta},o_{t+2\Delta},\ldots,o_{t+K\Delta}\right)
  \sim g_{\Theta}\left(\cdot|\mathcal O_t^h,s_t,l,c_t\right),
\end{equation}
where $K=\lfloor p/\Delta\rfloor$. In our implementation, $\Delta=3$ and $p=24$, so each inference chunk contains $K=8$ future visual frames and $24$ low-level action steps.

\begin{figure*}[!t]
\centering
\includegraphics[width=0.96\textwidth]{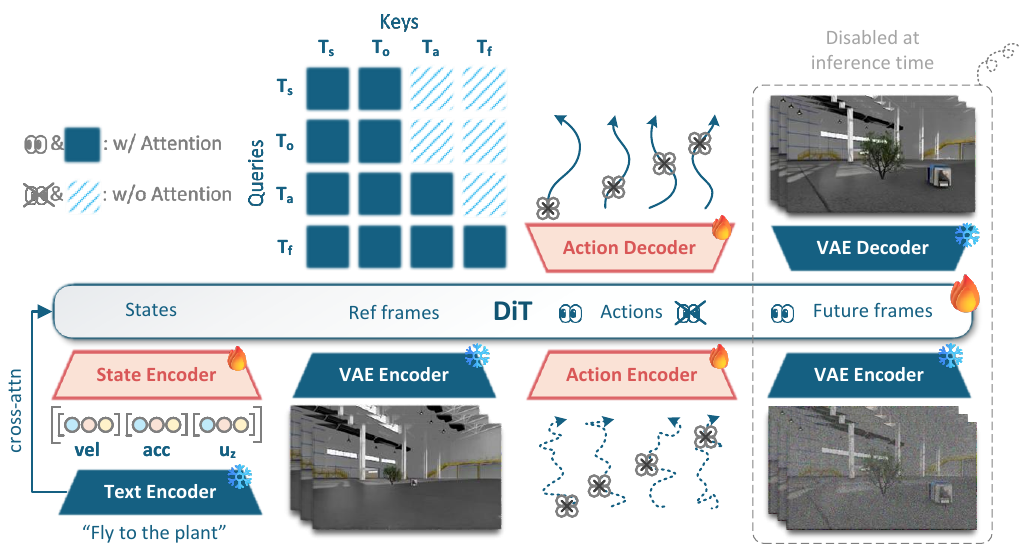}
\vspace{-0.cm}
\caption{\textbf{Action-centered world-action Transformer.} Visual history, proprioception, and language condition the action decoder. Future-frame tokens are used during training as consequence supervision, but are masked from action tokens and can be disabled at inference time.}\label{fig:architecture}
\vspace{-0.cm}
\end{figure*}

\subsection{Model Architecture}

\name{} adapts a 1.3B-parameter Wan video diffusion Transformer~\citep{wan} to aerial navigation. Encoding and predicting every $10$ Hz video frame is computationally expensive and visually redundant, so we encode and forecast one observation every $\Delta$ steps. At training time, a clip $o_{t-(h-1)\Delta:t+K\Delta}$ is encoded by the frozen video VAE into spatiotemporal latent tokens. After 3D RoPE positional encoding, the tokens are divided into reference tokens $T_o$ and future tokens $T_f$. The proprioceptive state $s_t$ and action chunk $a_{t:t+p-1}$ are projected by MLP encoders into state tokens $T_s$ and action tokens $T_a$. The instruction $l$ is encoded by the pretrained text encoder into language tokens $T_l$.

Instead of introducing separate transformer experts, we concatenate the non-language tokens into a single sequence $T_t=[T_s;T_o;T_a;T_f]$. Language tokens $T_l$ are provided as cross-attention context. A blockwise causal mask preserves the action-consequence ordering: state and reference tokens cannot attend to predicted tokens; action tokens attend only to state, reference and action tokens; future visual tokens attend to all tokens. Thus, action prediction cannot leak information from future frames, while future-frame reconstruction remains conditioned on the generated action chunk. At inference, the model omits $T_f$ and decodes only the action stream, reducing memory and computation and avoiding error propagation from imagined images to control. With $4$ denoising steps, disabling video prediction reduces inference time from $0.296$s to $0.184$s, saving $37.8$\% of time.

The training objective couples action prediction with visual consequence modeling. Let $\epsilon_a$ and $\epsilon_f$ denote the target noise or velocity fields for action and future-frame latent tokens under the diffusion or flow-matching parameterization. The model is trained with
\begin{equation}
\mathcal L(\Theta)=\mathcal L_{\mathrm{act}}+\lambda_{\mathrm{vis}}\mathcal L_{\mathrm{vis}},
\end{equation}
where
\begin{equation}
\mathcal L_{\mathrm{act}}=\mathbb E\left[\|g_\Theta^a(T_s,T_o,T_a,T_l)-\epsilon_a\|_2^2\right],\quad
\mathcal L_{\mathrm{vis}}=\mathbb E\left[\|g_\Theta^f(T_s,T_o,T_a,T_f,T_l)-\epsilon_f\|_2^2\right].
\end{equation}
The visual term is used to shape the representation and action prior during training; deployment relies on $g_\Theta^a$ alone. This separation is important for flight: the model benefits from consequence supervision, but the closed-loop controller is not forced to wait for or trust generated future frames.

\subsection{Data}

\begin{figure*}[!t]
\centering
\includegraphics[width=0.96\textwidth]{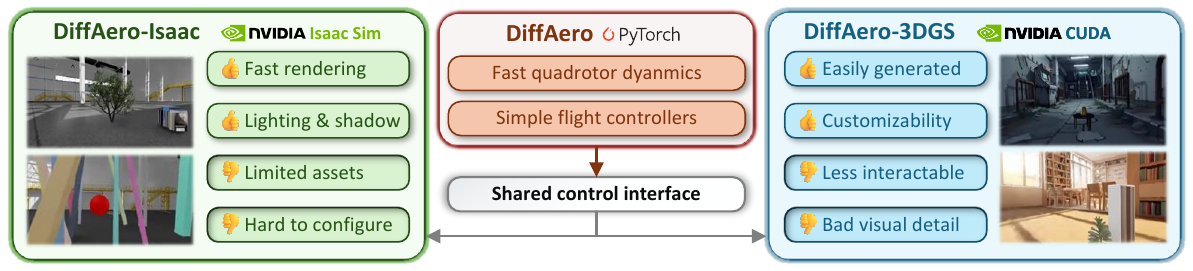}
\vspace{-0.cm}
\caption{\textbf{Simulation data pipeline.} DiffAero provides a shared dynamics and control interface. Isaac Lab improves visual fidelity, while the 3DGS branch scales scene and object diversity with lower asset-engineering cost.}\label{fig:simulators}
\vspace{-0.cm}
\end{figure*}

\paragraph{Simulators.} Ordinary video and embodied-agent datasets do not provide the synchronized supervision needed here: first-person RGB, proprioception, natural language, and action chunks must all be consistent with quadrotor motion. We therefore build the data stack around DiffAero~\citep{diffaero}, which supplies fast quadrotor dynamics and flight controllers. As shown in Fig.~\ref{fig:simulators}, two rendering branches share this control interface. The Isaac Lab branch provides realistic lighting, material, shadow, and camera effects, but scene diversity is limited by asset availability and manual configuration cost. The 3DGS branch renders RGB and depth directly from Gaussian-splat scenes, making it easier to scale across generated scenes and objects~\citep{hyworld2}; its limitation is lower fidelity for close-range geometry and lighting. The two branches are complementary: Isaac improves visual realism, while 3DGS improves scene and object diversity.

\paragraph{Scenarios.} We collect three classes of simulation demonstrations. First, in Isaac tracking, a target moves through random corridors and an expert controller keeps the quadrotor close to it, teaching motion anticipation and visual servoing. Second, in Isaac reaching, a language-specified object is placed among distractors and the quadrotor approaches it, linking language grounding to goal-directed flight. Third, in 3DGS reaching, a scene and object are sampled from the asset library, merged by placing the object on an empty plane, and used to collect object-approach demonstrations from randomized initial states. Together, these scenarios train the model to produce collision-aware local trajectories, ground language in egocentric observations, and forecast the visual consequences of predicted actions.

\begin{figure*}[htb]
\centering
\includegraphics[width=0.8\textwidth]{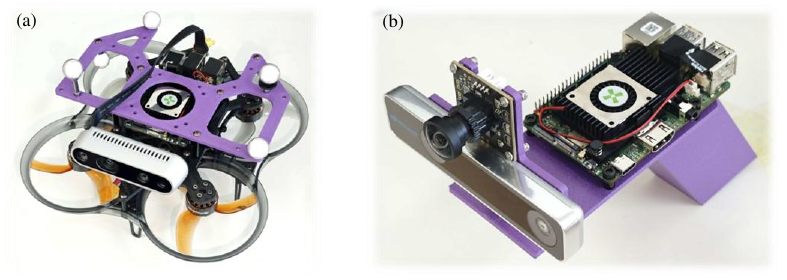}
\vspace{-0.2cm}
\caption{\textbf{Experimental hardware platforms.} (a) UAV platform for flight-task execution, equipped with onboard sensing modules and connected to a remote server via a local network for offboard inference. The resulting trajectories are tracked using OM-MPC. (b) Self-made portable data-collection platform for recording cropped camera observations and corresponding trajectory annotations, enabling efficient large-scale collection of interaction data.}
\label{fig:hardware}
\vspace{-0.cm}
\end{figure*}

\paragraph{Real-world data collection.} Directly collecting large-scale real quadrotor demonstrations remains expensive and risky. We therefore design a handheld device containing a fisheye camera, an Intel RealSense T265 stereo camera, and a lightweight computer, as shown in Fig.~\ref{fig:hardware}(b). The device estimates its own odometry with the T265 and converts the measured motion into simulated quadrotor poses. These poses are used to crop fisheye images into first-person views consistent with quadrotor camera motion. Using this setup, we collect $858$ real-world trajectories with $332,429$ frames, corresponding to approximately $3$ hours of data.

\begin{figure*}[!t]
\centering
\includegraphics[width=0.96\textwidth]{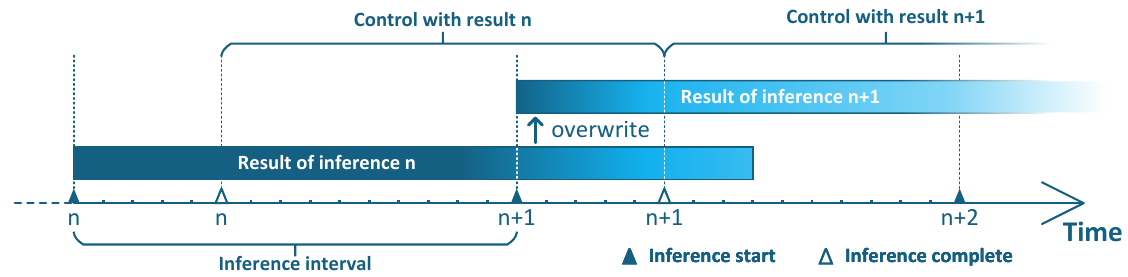}
\vspace{-0.cm}
\caption{\textbf{Inference-time self-guidance.} Consecutive action chunks overlap in time. The prefix of the new chunk is guided by the suffix of the previous chunk, reducing discontinuities before the remaining horizon is sampled.}\label{fig:selfguidance}
\vspace{-0.cm}
\end{figure*}

\subsection{Smooth Trajectory Generation with Self-Guidance}
Diffusion and flow-matching models~\citep{flowmatching} can represent multimodal action distributions, but independently sampled action chunks may differ across neighboring replanning calls. For quadrotors, such inconsistency can create jitter in the reference trajectory, especially when the action horizon is longer than the inference interval. We therefore introduce self-guidance to make consecutive action chunks agree over their overlapping time window.
Specifically, when inference $n+1$ starts before the chunk from inference $n$ has expired, the two chunks overlap, as illustrated in Fig.~\ref{fig:selfguidance}. During denoising, we replace the prefix velocity of the new sample with the corresponding suffix from the previous action chunk and sample only the remaining suffix in the usual way
\begin{subequations}
\begin{align}
    a_{n+1,0:\lfloor t_\text{infer}/t_\text{data}\rfloor}^{\sigma_i}&\leftarrow a_{n+1,0:\lfloor t_\text{infer}/t_\text{data}\rfloor}^{\sigma_{i-1}}+\delta t\times a_{n,\lfloor t_\text{interval}/t_\text{data}\rfloor:\lfloor (t_\text{interval}+t_\text{infer})/t_\text{data}\rfloor}^{1},\\
    a_{n+1,\lfloor t_\text{infer}/t_\text{data}\rfloor:p}^{\sigma_i}&\leftarrow a_{n+1,\lfloor t_\text{infer}/t_\text{data}\rfloor:p}^{\sigma_{i-1}}+\delta t\times g_{\Theta}\left(a_{n+1}^{\sigma_{i-1}}, o_{n+1},s_{n+1},l\right)_{\lfloor t_\text{infer}/t_\text{data}\rfloor:p},
\end{align}
\end{subequations}
where $t_\text{infer}$, $t_\text{interval}$, and $t_\text{data}$ denote inference latency, inference interval, and data time step, respectively, and $\delta t=\sigma_i-\sigma_{i-1}$. This operation does not change the model architecture; it constrains the sampler so that the low-level controller receives temporally consistent trajectory references.

\section{Experiments}
\label{sec:experiment}
\subsection{Implementation Details and Setup}
We use Wan2.1-1.3B~\citep{wan} as the backbone, including its video DiT, text encoder, and video VAE. The proprioception encoder, action encoder, and action decoder are MLPs with hidden sizes $128$ and $256$. Simulation and real-world demonstrations are collected at $30$ Hz and temporally downsampled by $3\times$, giving one video frame per three actions, eight future video frames, and $24$ actions per chunk. The model uses nine reference frames, corresponding to $2.4$ s of visual context.
Pretraining on simulation data takes $36$ hours on a single machine with $8\times$ NVIDIA A100 GPUs. We use AdamW~\citep{adamw} with learning rate $4.3\times10^{-5}$ and weight decay $0.01$. The pretraining set contains $900$K simulation video clips and $320$M frames: $500$K tracking clips, $200$K Isaac reaching clips, and $200$K 3DGS reaching clips.

All inference experiments are conducted on a single NVIDIA RTX 5090 24G Laptop GPU. The approximate VRAM usage is $4,500$ MB for both simulation experiments and real-world deployment. In real-world experiments, inference requests are sent to the workstation through ZeroMQ; the predicted action chunk is returned to the quadrotor, converted into local trajectory segments, and tracked by an on-manifold model predictive controller~\citep{ommpc}.

\subsection{Simulation Results}
We evaluate closed-loop performance in two Isaac Lab tasks: target tracking and language-conditioned object search. \name{}-FT denotes the model after real-world finetuning. The evaluation is designed to test whether the proposed temporal context and action-centered WAM design lead to executable trajectories rather than only accurate open-loop predictions.

\begin{table}[t]
  \centering
    \small
  \begin{tabular}{l|lc|cc}
    \toprule
     & Task & Reference frames& Success $\uparrow$ & Collision $\downarrow$ \\
    \midrule
    \name{}& tracking& 1& 20.0&90.0\\
    \name{}  & searching&1& 90.0 & 10.0 \\
    \midrule
    \name{}& tracking& 9& \textbf{100.0} & \textbf{0.0} \\
    \name{}  & searching&9& \textbf{100.0} & \textbf{0.0} \\
    \midrule
    \name{}-FT& tracking& 9& 95.0&\textbf{0.0}\\
    \name{}-FT  & searching&9& \textbf{100.0} & \textbf{0.0} \\
    \midrule
  \end{tabular}
    \vspace{0.2cm}
    \caption{\textbf{Closed-loop simulation results in Isaac Lab over $20$ episodes.} Success denotes target visibility for at least $80\%$ of the trajectory in tracking and target discovery in searching; collision denotes the percentage of episodes ending in collision.}
  \label{tab:main-results}
        \vspace{-0.cm}
\end{table}
\begin{table}[t]
  \centering
    \small
  \begin{tabular}{l|c|ccc}
    \toprule
     Model& Reference frames& Average final distance & Success $\uparrow$ & Collision $\downarrow$ \\
    \midrule
    \name{} & 1& \textbf{3.819}&\textbf{75.0}&\textbf{25.0}\\
    \midrule
    \name{} & 9& 1.983&\textbf{100.0} & \textbf{0.0} \\
    \midrule
    \name{}-FT & 9&1.988& \textbf{100.0}&\textbf{0.0}\\
    \midrule
  \end{tabular}
    \vspace{0.2cm}
    \caption{Simulation results on the searching task with unseen target objects.}
  \label{tab:ood-results}
      \vspace{-0.5cm}
\end{table}

With only one reference frame, the model struggles in closed-loop tracking, achieving $20\%$ success and $90\%$ collision. This confirms that a single egocentric image is insufficient for estimating relative motion and producing safe trajectory chunks. Using nine reference frames improves both tracking and searching to $100\%$ success with zero collisions. The improvement also holds for unseen target objects, where success increases from $75\%$ to $100\%$ and the average final distance decreases from $3.819$ m to $1.983$ m. After real-world finetuning, \name{}-FT maintains strong simulation performance with zero collisions and only minor success degradation, suggesting that real-world adaptation does not substantially compromise the learned closed-loop policy.

\subsection{Real-World Flight}
\begin{figure*}[!t]
\centering
\includegraphics[width=0.96\textwidth]{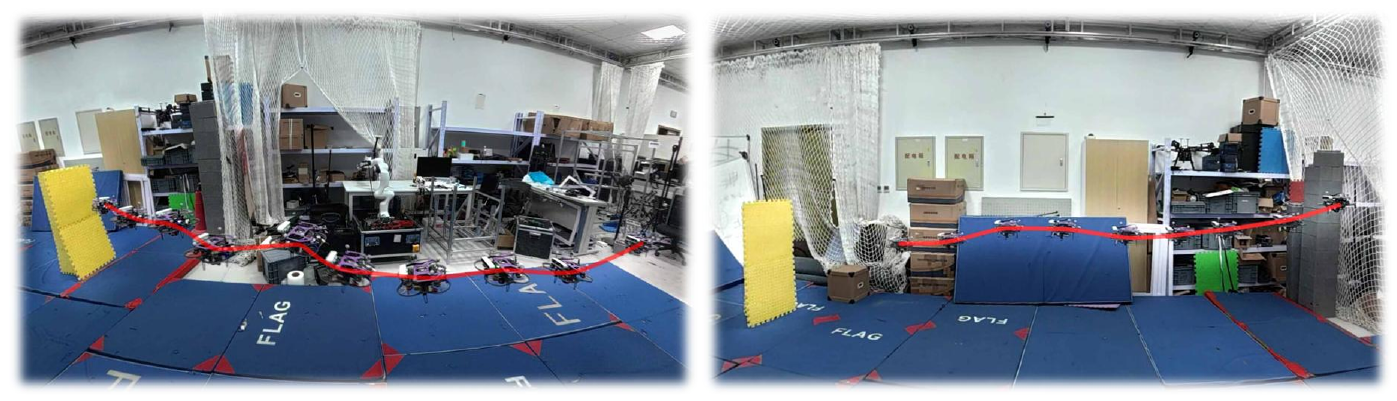}
\vspace{-0.cm}
\caption{\textbf{Real-world flight with language-conditioned WAM.} AeroAct generates feasible flight commands and guides the quadrotor toward the specified target object in a lab environment. The trajectory illustrates closed-loop execution of the learned WAM policy on a physical aerial robot.}
\label{fig:realworld}
\vspace{-0.cm}
\end{figure*}

We further validate \name{} through real-world flight experiments. The quadrotor platform is built by modifying an OddityRC 35Pro frame and is equipped with an Intel RealSense D435i camera and a Radxa ROCK 5C onboard computer. The experiments are conducted in a laboratory environment with a motion-capture system for state measurement and safety monitoring. Under this hardware configuration, the inference latency of the model is approximately $t_{\mathrm{infer}}=0.8$~s.

To the best of our knowledge, this is the first WAM-based real-world quadrotor flight demonstration, where a learned model directly connects visual-language task specification, predictive action generation, and closed-loop quadrotor execution. In the experiment shown in Fig.~\ref{fig:realworld}, the instruction given to the model is ``fly to the yellow foam mat.'' The model successfully generates feasible flight commands and navigates the quadrotor toward the target object, demonstrating the potential of world-action modeling for embodied aerial intelligence beyond simulation.

\section{Limitations}
\label{sec:limitations}
Our real-world experiments are currently limited to short indoor trajectories. The temporal context is sufficient for single-stage target reaching, but not yet for complex instructions that require multiple semantic sub-goals, recovery behaviors, or long-horizon memory. Real-world deployment also relies on offboard inference because of the computational cost of the video-diffusion backbone.

\section{Conclusion}
\label{sec:conclusion}
We presented \name{}, an action-centered WAM for language-conditioned quadrotor flight. Instead of predicting discrete actions, high-level waypoints, or instantaneous velocities, \name{} outputs smooth local trajectory chunks and uses future egocentric visual prediction as training-time consequence supervision. This design preserves the temporal and geometric priors of generative video models while enabling efficient action decoding without test-time video generation.

To support aerial WAM training, \name{} combines DiffAero dynamics, Isaac Lab, 3DGS rendering, and handheld real-world demonstrations into a hybrid data pipeline with aligned vision, language, proprioception, and dynamically feasible actions. Simulation results show that temporal visual history and mixed-domain data improve target tracking and language-conditioned object search. Real-world experiments further validate, to the best of our knowledge, the first WAM-based quadrotor flight on a physical platform. Current limitations include short indoor trials and reliance on offboard inference. Future work will focus on onboard efficiency, robustness under more aggressive flight dynamics, uncertainty-aware action generation, and long-horizon multi-stage aerial instruction following.



\acknowledgments{This work was supported in part by the National Natural Science Foundation of China under Grants U23B2059, 62088101, and also by the Zhongguancun Academy under Grant 20240307.}


\bibliography{example}  

\newpage
\appendix
\section{Appendix}

\subsection{Observation and Action Spaces}

Proprioception and actions are expressed in a yaw-aligned local frame. Its $x$-axis lies in the vertical plane defined by the body $x$-axis, and its $z$-axis points upward. The proprioceptive input is
\begin{equation}
    s=[\mathbf{v}^{l\top},\mathbf{a}^{l\top},\mathbf{u}_{\mathrm z}^{l\top}]^\top\in\mathbb R^9,
\end{equation}
where $\mathbf v^l$, $\mathbf a^l$, and $\mathbf u_\mathrm z^l$ are the local velocity, local acceleration, and body-$z$ unit vector represented in the local frame.

Each action parameterizes a local trajectory segment rather than an instantaneous velocity command. For each dimension $\mu\in\{x,y,z\}$, the segment is a fifth-order polynomial
\begin{equation}
  \mathbf{p}^l_\mu(t)=\alpha_0 + \alpha_1 t + \alpha_2 t^2 + \alpha_3 t^3 + \alpha_4 t^4 + \alpha_5 t^5.
\end{equation}
Similar to \citet{yopo}, the action specifies the endpoint of a $T=2$ s local trajectory segment
\begin{equation}
    a=[r, \theta, \psi, \mathbf v_\text{end}^{l\top}, \mathbf a_\text{end}^{l\top}]^\top\in\mathbb R^9,
\end{equation}
where $\mathbf v_\text{end}^{l}$ and $\mathbf a_\text{end}^{l}$ are endpoint velocity and acceleration, and $(r,\theta,\psi)$ defines the endpoint displacement by
\begin{equation}
  \left\{
    \begin{aligned}
      p_{x,\text{end}}^l &= r \cos(\theta) \cos(\psi), \\
      p_{y,\text{end}}^l &= r \cos(\theta) \sin(\psi), \\
      p_{z,\text{end}}^l &= -r \sin(\theta).
    \end{aligned}
  \right.
\end{equation}
Given $\mathbf p^l_\text{start}=\mathbf 0$, $\mathbf v_\text{start}^l$, $\mathbf a_\text{start}^l$, and the endpoint conditions, the polynomial coefficients for $\mu\in\{x,y,z\}$ are obtained as follows
\begin{equation}
  \begin{bmatrix}
    \alpha_0^\mu \\
    \alpha_1^\mu \\
    \alpha_2^\mu \\
    \alpha_3^\mu \\
    \alpha_4^\mu \\
    \alpha_5^\mu
  \end{bmatrix}=
  \begin{bmatrix}
    1 & 0 & 0 & 0 & 0 & 0 \\
    0 & 1 & 0 & 0 & 0 & 0 \\
    0 & 0 & 2 & 0 & 0 & 0 \\
    1 & T & T^2 & T^3 & T^4 & T^5 \\
    0 & 1 & 2T & 3T^2 & 4T^3 & 5T^4 \\
    0 & 0 & 2 & 6T & 12T^2 & 20T^3
  \end{bmatrix}^{-1}\begin{bmatrix}
    p_{\mu,\text{start}}^l \\
    v_{\mu,\text{start}}^l \\
    a_{\mu,\text{start}}^l \\
    p_{\mu,\text{end}}^l \\
    v_{\mu,\text{end}}^l \\
    a_{\mu,\text{end}}^l
  \end{bmatrix}.
\end{equation}

\subsection{Evaluation Metrics}
\paragraph{Tracking metrics.} Tracking evaluates whether the UAV follows a saved moving-target trajectory. The final-distance metric is
\begin{equation}
    \mathrm{dist\_m}_i=d_i^{\mathrm{final}}=d_i(T_i),\quad
    D_{\mathrm{final}}^{\mathrm{trk}}=
    \frac{1}{N}\sum_{i=1}^{N}\mathrm{dist\_m}_i .
\end{equation}
Because final distance alone does not certify that the target stayed visible, we also report collision and camera-based metrics
\begin{equation}
R_{\mathrm{coll}}^{\mathrm{trk}}=
    \frac{1}{N}\sum_{i=1}^{N}
    \mathbb I\left(\exists t,\ \mathrm{terminated}_i(t)\vee
    \mathrm{collision}_i(t)\right).
\end{equation}
Let $v_i(t)$ indicate whether the target center is inside the camera frustum, and let $c_i(t)$ indicate whether its normalized image-plane offset is no larger than $\tau_c=0.25$:
\begin{equation}
    R_{\mathrm{view},i}=
    \frac{1}{K_i}\sum_{t=0}^{K_i-1}v_i(t),\quad
    R_{\mathrm{center},i}=
    \frac{1}{K_i}\sum_{t=0}^{K_i-1}c_i(t),
\end{equation}
with dataset-level averages
\begin{equation}
    R_{\mathrm{view}}=\frac{1}{N}\sum_{i=1}^{N}R_{\mathrm{view},i},\quad
    R_{\mathrm{center}}=\frac{1}{N}\sum_{i=1}^{N}R_{\mathrm{center},i}.
\end{equation}
The centered-target rate is stricter than the in-view rate because it additionally requires the target to remain close to the optical axis.

\paragraph{Searching metrics.} The searching task measures whether the language-conditioned policy reaches a specified object without collision. An episode reaches the target if the environment reports success or the drone enters the $\tau_s=2.0$ m success radius at any time. The success metric additionally requires no collision
\begin{equation}
R_{\mathrm{succ}}^{\mathrm{srch}}=
    \frac{1}{N}\sum_{i=1}^{N}
    \mathbb I\left(
\left(\mathrm{success}^{\mathrm{env}}_i\vee
    \exists t,\ d_i(t)\le \tau_s\right)
    \wedge \neg \mathrm{collided}_i
    \right).
\end{equation}
The collision rate is
\begin{equation}
    R_{\mathrm{coll}}^{\mathrm{srch}}=
    \frac{1}{N}\sum_{i=1}^{N}\mathbb I\left(
    \exists t,\ \mathrm{terminated}_i(t)\vee \mathrm{collision}_i(t)\right),
\end{equation}
and the final distance is
\begin{equation}
\mathrm{dist\_m}_i=\mathrm{final\_dist\_m}_i=d_i(T_i),\quad
    D_{\mathrm{final}}^{\mathrm{srch}}=
    \frac{1}{N}\sum_{i=1}^{N}\mathrm{dist\_m}_i .
\end{equation}
In the current logging code, \texttt{dist\_m} and \texttt{final\_dist\_m} denote the same final-distance value.

\subsection{Additional Experimental Results}
Let $\mathbf p_i(t)$ be the drone position in rollout $i$ at simulation time $t$, and let $\mathbf g_i(t)$ denote the target position. The instantaneous target distance and final valid step are
\begin{equation}
    d_i(t)=\left\|\mathbf p_i(t)-\mathbf g_i(t)\right\|_2,\quad
    T_i=K_i-1 .
\end{equation}
All reported means are averaged over $N$ trajectories or episodes. The policy is queried every
\begin{equation}
n_{\mathrm{replan}}=\max\!\left(1,
    \operatorname{round}\!\left(\tfrac{1}{f_{\mathrm{server}}dt}\right)\right)
\end{equation}
simulation steps, where $f_{\mathrm{server}}$ is the inference frequency and $dt$ is the simulation step size.

We compare three checkpoints under the same Isaac Lab scenes, camera model, controller settings, and inference-server interface. Tracking uses $20$ trajectories at $1$, $2$, and $5$ Hz. Searching uses $20$ language-specified object-search episodes at $1$, $2$, and $5$ Hz. For tracking, final distance can be misleading: a rollout may end near the target after losing it, or may keep the target in view while ending farther away. We therefore treat target visibility as the primary tracking signal. In Table~\ref{tab:tracking-results}, view success counts a trajectory as successful when its target-in-view rate is at least $80\%$. Table~\ref{tab:searching-results} reports success, collision and final target distance.
\begin{table}[h]
  \centering
  \setlength{\tabcolsep}{2.5pt}
  \begin{tabular}{l|cc|ccccc}
    \toprule
    Model&Reference frames& Freq. & View succ. $\uparrow$ & In-view $\uparrow$ & Centered $\uparrow$ & Dist. $\downarrow$ & Coll. $\downarrow$ \\
    \midrule
    \name{}  &1& 1 Hz & 20.0 & 55.6 & 21.8 & 7.902 & 90.0 \\
    \name{}  &1& 2 Hz & 5.0 & 55.7 & 25.3 & 12.306 & 100.0 \\
    \name{}  &1& 5 Hz & 10.0 & 57.9 & 30.9 & 13.163 & 90.0 \\
    \midrule
    \name{}  &9& 1 Hz & \textbf{100.0} & \textbf{100.0} & 64.2 & 3.436 & \textbf{0.0} \\
    \name{}  &9& 2 Hz & \textbf{100.0} & 99.3 & 77.0 & \textbf{2.245} & \textbf{0.0} \\
    \name{}  &9& 5 Hz & \textbf{100.0} & 98.7 & 78.7 & 2.375 & 5.0 \\
    \midrule
    \name{}-FT  &9& 1 Hz & 95.0 & 97.5 & 61.0 & 2.845 & \textbf{0.0} \\
    \name{}-FT  &9& 2 Hz & \textbf{100.0} & 99.2 & 73.3 & 2.865 & \textbf{0.0} \\
    \name{}-FT  &9& 5 Hz & \textbf{100.0} & 99.4 & \textbf{81.8} & 2.668 & \textbf{0.0} \\
    \bottomrule
  \end{tabular}
  \vspace{0.4cm}
  \caption{Closed-loop tracking results in simulation. View success is the percentage of $20$ trajectories whose target-in-view rate is at least $80\%$. In-view, centered, and collision are percentages; distance is measured in meters.}
  \label{tab:tracking-results}
\end{table}

\begin{table}[h]
  \centering
  \begin{tabular}{l|cc|ccc}
    \toprule
    Model&Reference frames& Freq. & Success $\uparrow$ & Collision $\downarrow$ & Final dist. $\downarrow$ \\
    \midrule
    \name{}  &1& 1 Hz & 90.0 & 10.0 & 2.616 \\
    \name{}  &1& 2 Hz & 80.0 & 20.0 & 3.421 \\
    \name{}  &1& 5 Hz & 75.0 & 20.0 & 4.064 \\
    \midrule
    \name{}  &9& 1 Hz & \textbf{100.0} & \textbf{0.0} & \textbf{1.982} \\
    \name{}  &9& 2 Hz & \textbf{100.0} & \textbf{0.0} & \textbf{1.980} \\
    \name{}  &9& 5 Hz & \textbf{100.0} & \textbf{0.0} & 1.981 \\
    \midrule
    \name{}-FT  &9& 1 Hz & \textbf{100.0} & \textbf{0.0} & \textbf{1.984} \\
    \name{}-FT  &9& 2 Hz & \textbf{100.0} & \textbf{0.0} & 1.987 \\
    \name{}-FT  &9& 5 Hz & \textbf{100.0} & \textbf{0.0} & \textbf{1.984} \\
    \bottomrule
  \end{tabular}
    \vspace{0.2cm}
  \caption{Closed-loop searching results in simulation. Success and collision are percentages over $20$ episodes, and final distance is measured in meters.}
  \label{tab:searching-results}
\end{table}

\end{document}